\documentclass[letterpaper, 10 pt, conference]{ieeeconf}  

\IEEEoverridecommandlockouts                              

\usepackage{amsmath} 
\usepackage{amssymb}  
\usepackage{graphicx}

\title{\LARGE \bf
Semi-on-Demand Transit Feeders with Shared Autonomous Vehicles and Reinforcement-Learning-Based Zonal Dispatching Control
}

\author{Max T.M. Ng$^{1}$, Roman Engelhardt$^{2,*}$, Florian Dandl$^{2}$, Hani S. Mahmassani$^{1}$, and Klaus Bogenberger$^{2}$
\thanks{This work is based on research funded in part by the German Academic Exchange Service and Northwestern Buffett Institute for Global Affairs to the first author. Additionally, this work is partially funded by the German Federal Ministry of Education and Research via the MCube Project STEAM.}
\thanks{$^{1}$Northwestern University Transportation Center, USA}%
\thanks{$^{2}$Chair of Traffic Engineering and Control, Technical University of Munich, Germany}%
\thanks{$^{*}$Corresponding author: {\tt\small roman.engelhardt@tum.de}}%
}

\begin{document}

\maketitle
\thispagestyle{empty}
\pagestyle{empty}

\begin{abstract}
This paper develops a semi-on-demand transit feeder service using shared autonomous vehicles (SAVs) and zonal dispatching control based on reinforcement learning (RL). 
This service combines the cost-effectiveness of fixed-route transit with the adaptability of demand-responsive transport to improve accessibility in lower-density areas. 
Departing from the terminus, SAVs first make scheduled fixed stops, then offer on-demand pick-ups and drop-offs in a pre-determined flexible-route area.
Our deep RL model dynamically assigns vehicles to subdivided flexible-route zones in response to real-time demand fluctuations and operations, using a policy gradient algorithm --- Proximal Policy Optimization.
The methodology is demonstrated through agent-based simulations on a real-world bus route in Munich, Germany. 
Results show that after efficient training of the RL model, the semi-on-demand service with dynamic zonal control serves 16\% more passengers at 13\% higher generalized costs on average compared to traditional fixed-route service. 
The efficiency gain brought by RL control brings 2.4\% more passengers at 1.4\% higher costs.
This study not only showcases the potential of integrating SAV feeders and machine learning techniques into public transit, but also sets the groundwork for further innovations in addressing first-mile-last-mile problems in multimodal transit systems.

\end{abstract}

\section{Introduction}

\begin{figure}[thpb]
      \centering
      \framebox{\parbox{3.3in}{
      \centering
      \includegraphics[width=3.3in]{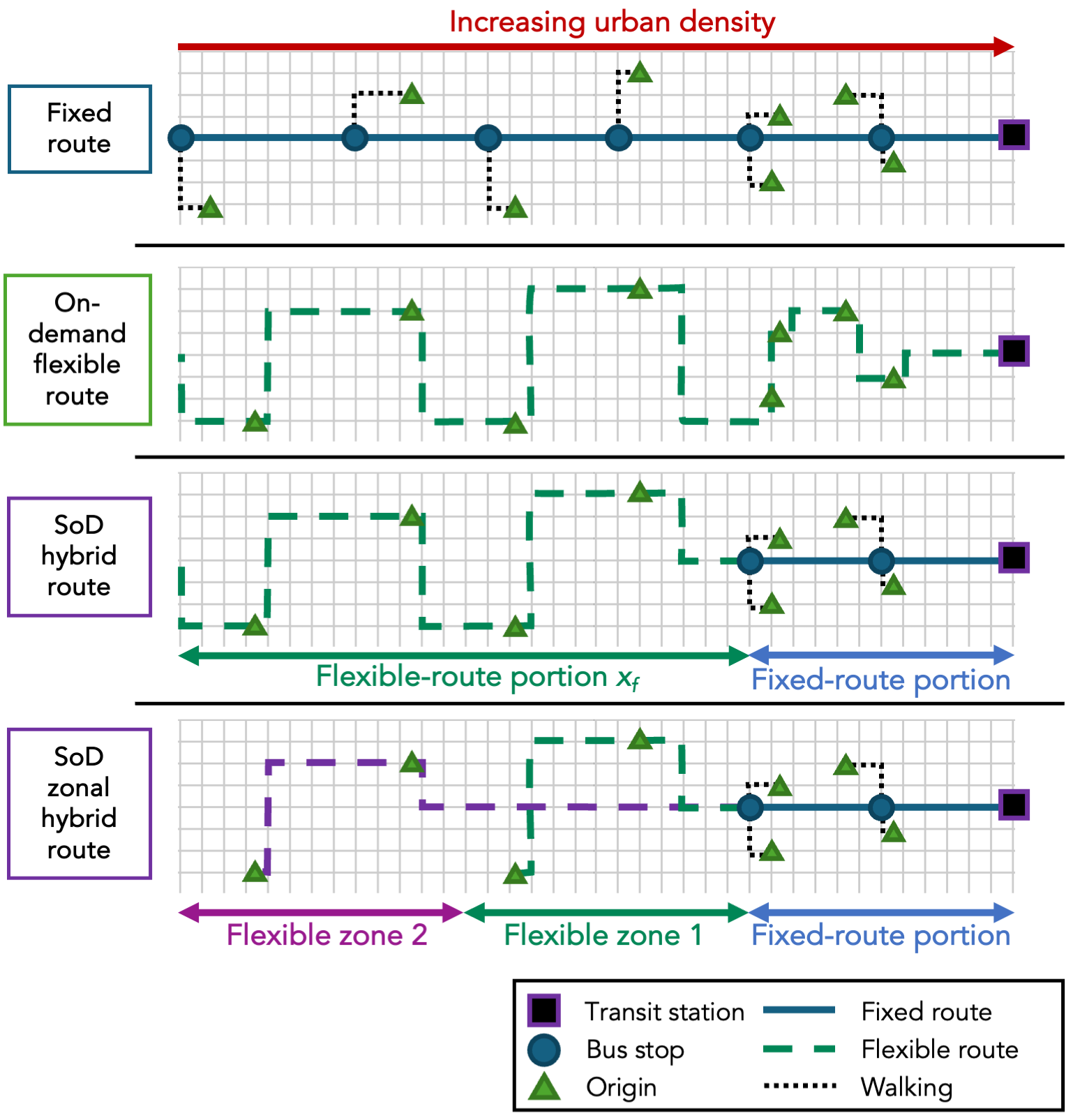}
      }}
    \caption{Illustration of various routes as a feeder service (SoD: semi-on-demand)}
    \label{fig:route_forms}
    \end{figure}

The advent of shared autonomous vehicles (SAVs) presents new opportunities to enhance multimodal public transit systems \cite{salazar_intermodal_2020, zhao_enhanced_2022}, for example, as feeders or demand-responsive transit (DRT) \cite{vansteenwegen_survey_2022, ng_redesigning_2024}. 
The SAV capabilities of central coordination enable more sophisticated route concepts, such as semi-on-demand (SoD) hybrid routes, which combine the economy of scale of fixed-route buses and the flexibility of DRT \cite{ng_autonomous_2023} (Fig.~\ref{fig:route_forms}). 
As a scheduled service, SAVs depart from the terminus, first make regular stops in the fixed-route area (higher demand density), then offer on-demand pick-ups and drop-offs akin to ride-pooling in the flexible-route area (lower demand density), and return to the fixed-route scheduled stops and the terminus. 
The demarcation of fixed-route and flexible-route areas is pre-determined at a service planning level for regular service patterns that are clear to passengers (whether to walk to a fixed stop or wait to be picked up).
SoD services reduce access times and provide passengers with flexibility and schedule predictability \cite{errico_survey_2013}, whereas such convenience is expected to attract more demand \cite{frei_flexing_2017}.

Nevertheless, flexible routing often suffers inefficiencies brought by excessive detours that degrade service quality and increase operating costs \cite{ng_semi--demand_2024}. 
This study addresses this problem by dividing the flexible service area into zones (similar to previous works on a service planning level \cite{li_2-vehicle_2011, lee_optimal_2024}). 
On top of regular frequency, special trips to a zone are dispatched with dynamic control to serve most passengers. 
This is supported by reinforcement learning (RL) techniques, widely used in ride-hailing and SAV dispatching (e.g., \cite{gueriau_samod_2018, mao_dispatch_2020, liu_context-aware_2022, liu_deep_2022}; see reviews \cite{qin_reinforcement_2022,haydari_deep_2022}), as well as customized bus planning (e.g., \cite{li_multiline_2024, wu_multi-agent_2024}). 
However, limited research has utilized RL in DRT control. In this study, we adopt a deep RL policy gradient algorithm --- Proximal Policy Optimization (PPO) \cite{schulman_proximal_2017} --- for its solution quality and performance (e.g., support for parallel training). 

This study aims to develop a SoD zonal vehicle dispatching strategy that dynamically aligns with varying demand and vehicle availability to maximize the number of passengers served, while ensuring frequent service along the key corridor (fixed-route portion) with base SoD route frequency.
We focus on scenarios of directional demand (e.g., feeders) in a relatively narrow corridor and off-peak hours, utilizing spare SAVs that serve fixed routes in peak hours to provide more flexible services.

Our contributions are two-fold. 
First, we introduce a novel SoD transit feeder service with SAVs and dynamic dispatching control. 
It features the economy of scales and schedule predictability of fixed routes and flexibility of flexible routes without excessive detours or waiting times. 
Second, we model the optimization problem of SAV zonal dispatching as a Markov decision process and develop an efficient RL-based SAV zonal dispatching control strategy with a policy gradient method. 
Its practical application is demonstrated with agent-based simulations and real-world demand data in Munich, Germany.

\section{Methodology}

\begin{figure}[thpb]
      \centering
      \framebox{\parbox{3.3in}{
      \centering
      \includegraphics[scale=0.095]{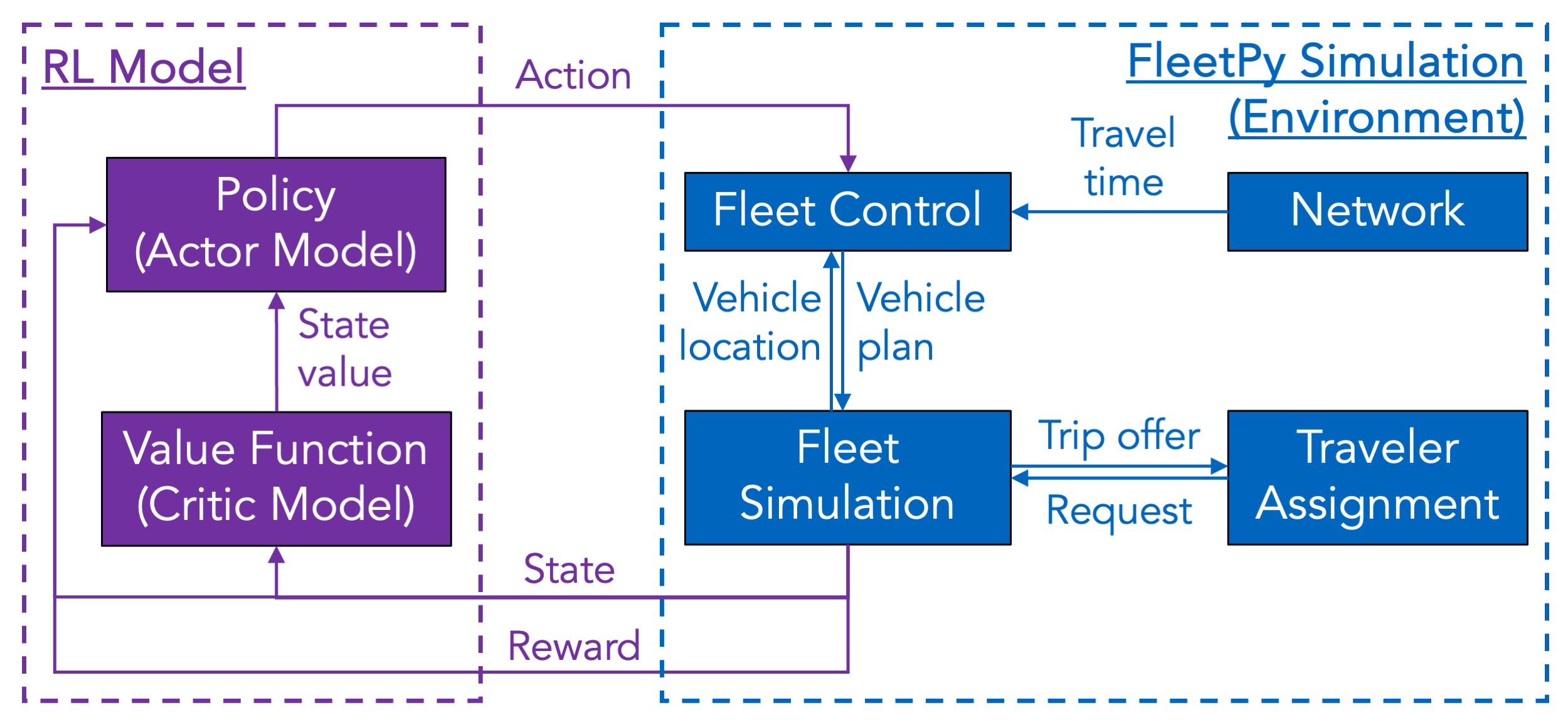}
      }}
    \caption{Components and data flows between the RL model and simulation}
    \label{fig:rl_fleetpy}
    \end{figure}

The dynamic dispatching control framework is built on FleetPy \cite{engelhardt_fleetpy_2022}, an agent-based SAV simulation framework adapted to semi-on-demand scheduled feeder services \cite{ng_simulating_2024}.
Fig.~\ref{fig:rl_fleetpy} summarizes the interaction between the RL model and simulation environment.
\footnote{The loops of RL model and simulation do not necessarily run on same frequency, e.g., RL model is updated every 5 minutes but simulation is updated every minute.}

\subsection{SAV Fleet Simulation and Control}

This simulation\footnote{Refer to \cite{engelhardt_fleetpy_2022} for simulation details using FleetPy.}
involves three agents: customers, a service operator, and an SAV fleet. 
Customers request trips $r \in \mathcal{R}$, each defined by a request time and origin-destination pair in a street network $G(N,E)$. 
The operator assigns requests to SAVs $v \in \mathcal{V}$, which follow a pre-set schedule of fixed stops and time reserved for the flexible-route portion $t^R$.
\footnote{The following discussion focuses on trips from the terminus for simplicity, but the framework also covers the opposite direction to the terminus and pick-ups/drop-offs along the way.}
\footnote{All superscripts are qualifiers and all subscripts are indices.}

In each time step $k$ (of size $t^S$) during simulation, the operator iteratively processes each request by matching it to the closest fixed stop (if in the fixed-route portion) or inserting a new stop into a vehicle schedule $\psi$ (if in the flexible-route portion).
Insertion heuristics with an exhaustive search compute all feasible insertions for a request and each vehicle to minimize $\rho(\psi)$ in \eqref{eq:rho} that weighs vehicle distance $d_{v,\psi}$, traveling time $a_{r,\psi}-t_r$, requests satisfied $n^R_{\psi}$, and requests served at fixed stops $n^{S}_{\psi}$ with respective cost coefficients $\gamma^O$, $\gamma^T$, $\gamma^R$, and $\gamma^S$:
    \begin{align}
        \rho(\psi) 
        = \sum_{v \in \mathcal{V}} \gamma^{O}_v d_{v,\psi} 
        + \sum_{r \in \mathcal{R}} \gamma^{T} (a_{r,\psi}-t_r) 
        - \gamma^{R} n^{R}_{\psi} 
        - \gamma^{S} n^S_{\psi} 
        \label{eq:rho}
    \end{align}
while respecting constraints on maximum waiting time $t^{W,max}$, maximum in-vehicle travel time (as factor $\phi^P$ of the shortest path time $t^D$ plus a constant $t^{D,f}$), and vehicle capacity $c^V$. 
In case of no feasible insertion found, the request is rejected.
With the assigned schedule, SAVs then travel between stops on the shortest paths.

The total generalized cost $c^G$ is calculated considering both user and operator perspectives in \eqref{eq:c^G}. 
The first term reflects the total users' costs, where $t^A_r$, $t^W_r$, and $t^T_r$ are the access\footnote{This study assumes walking as the access mode, but other active mobility modes, e.g., cycling and scooters can be readily considered.}, waiting, and riding times for a specific request $r$, and $\gamma^A$, $\gamma^W$, and $\gamma^T$ are the cost coefficients, respectively. 
The second term is the total operating cost, with $d_v$ and $t^V_v$ as the distance traveled and time deployed of vehicle $v$, and $\gamma^O$ (per distance), and $\gamma^V$ (per vehicle-hour) the respective cost coefficients.

    \begin{align}
        c^G &= \sum_{r \in \mathcal{R}}{(\gamma^A t^A_r + \gamma^W t^W_r + \gamma^T t^T_r)} + \sum_{v \in \mathcal{V}}{(\gamma^O_v d_v + \gamma^V_v t^V_v)}
        \label{eq:c^G}
    \end{align}

For zonal control, each SAV $v$ in each cycle is assigned a zone ($z_v=1, 2$) or regular route (i.e., all zones; $z_v=0$). 
Following SoD operations, it first traverses the fixed-route portion, then picks up/drops off passengers in the designated flexible zones, and finally, returns to the fixed-route portion and terminus.
Passengers are assigned to any SAVs that serve their origin-destination zones. For example, passengers traveling between the terminus and fixed-route portion can take any SAV, but passengers traveling between the terminus and Zone 2 can only take SAV $v|z_v=0 \text{ or } 2$ but not those assigned to Zone 1.

\subsection{Reinforcement Learning}

The optimization problem to maximize passengers served by dispatching SAVs from the terminus at each time step $k$ is modeled as a Markov decision process and solved with an RL model.
There are four possible actions $a$ --- 1) assign and send an SAV to regular SoD service ($a_k=0$); 2\&3) assign and send an SAV to Zone 1 ($a_k=1$) or 2 ($a_k=2$); 4) and do not send an SAV ($a_k=3$).
An action is only taken if an SAV is available at the terminus. 
Besides, an SAV from a reserved fleet is sent to regular SoD service per a constant period to maintain minimum service that overrides the RL model.\footnote{Alternative RL models may consider invalid actions through penalty or masking \cite{huang_closer_2022}, but preliminary testing found no considerable performance improvement.} 
After a dispatching decision, an SAV would start boarding and leave the terminus in 5~min.

To represent the system state $s$ parsimoniously while confirming with the Markov property, 18 state variables are selected: numbers of running and available SAVs, 15-min demand forecast, and for each zone, number of unassigned requests, scheduled SAV flexible time, number of assigned boarding/alighting processes, and time since the last SAV departure. 
Each state variable is normalized with its perceived value range.

The reward $r$ is set as the negative number of request rejections due to prolonged waiting time.\footnote{Using request rejections as the reward function allows easy estimation of state values by the critic model. Future works may consider generalized costs that incorporate waiting time and journey time alongside request satisfaction.}

This study employs PPO, a policy gradient method that trains a stochastic policy $\pi$ in an on-policy manner, for its support of discrete actions\footnote{Deep Q Network is an alternative RL method that also supports discrete actions.} and parallel training.

The actor-critic framework (see Fig.~\ref{fig:rl_fleetpy}) involves an actor (policy) proposing actions based on the state, and a critic (value function estimator) evaluating these actions.
The impacts of SAV dispatching at different time steps are intertwined and realized only after some time (after SAVs are sent to a zone and pick up passengers). 
Therefore, the critic model aids in estimating state values and then the advantage function $A_{\pi_\theta}(s,a)$, which is the expected benefits brought by an action $a$ over any action in state $s$. 

For exploration, actions are sampled according to the latest policy $\pi_{\theta_k}(a|s)$ of the actor model, which returns a probability of choosing action $a$ given state $s$. 
The knowledge of advantage provided by the critic model allows the exploitation of more certain rewards.

During training, the states, actions, and rewards of multiple time steps (i.e., the SAV dispatching and simulation results) are used to train the critic model in each episode. 
It uses Generalized Advantage Estimation (GAE) \cite{schulman_high-dimensional_2018} in \eqref{eq:GAE} that balances bias and variance\footnote{Including more future estimates decreases bias but increases variance.} by discounting future advantage estimates $\delta_{t+k}$ with a factor of $\lambda$ (future rewards $r_t$ are discounted separately with $\gamma$):
    \begin{align}
        A_t &= \sum_{k=0}^{T-t} (\gamma \lambda)^k \delta_{t+k}
        \label{eq:GAE}
    \end{align}
where $T$ is the batch size, and each $\delta_t$ is the temporal difference error of a single time step with reference to the value function $V(s_t)$ in \eqref{eq:advantage}:
    \begin{align}
        \delta_t &= r_t + \gamma V(s_{t+1}) - V(s_t)
        \label{eq:advantage}
    \end{align}

In all optimization steps within the same episode, the actor model adjusts policy parameters $\theta$ to maximize expected returns in \eqref{eq:theta_k}:
    \begin{align}
        \theta_{k+1} &= \text{argmax}_\theta \mathbf{E}_{s,a \sim \pi_{\theta_k}} \left[L(s,a,\theta_k,\theta)\right]
        \label{eq:theta_k}
    \end{align}
$L(s,a,\theta_k,\theta)$ is the loss function in \eqref{eq:L}, which is the surrogate objective function (ratios of current and old policy to take action $a$ given state $s$, multiplied by the advantage estimate from the critic model) clipped by $\epsilon$ in \eqref{eq:clip}.
        \begin{align}
        L(s,a,\theta_k,\theta) &= \min \left\{
        \begin{array}{c}
        \frac{\pi_\theta(a|s)}{\pi_{\theta_k}(a|s)} A_{\pi_{\theta_k}}(s,a) \\
        g\left(\epsilon,  A_{\pi_{\theta_k}}(s,a)\right)
        \end{array}
        \right\}
        \label{eq:L}
        \\
        g(\epsilon,  A) &=
        \begin{cases}
            (1+\epsilon) A, A \geq 0 \\
            (1-\epsilon) A, A < 0
        \end{cases}
        \label{eq:clip}
    \end{align}
This clipping mechanism leads to the stability of PPO by limiting the extent of policy changes.
 
The policy network architecture for both actor and critic models consists of two-layer fully connected networks\footnote{Each layer has 64 units. The activation function is tanh, with softmax (actor) and linear (critic) for an extra output layer. Parameters are optimized with an Adam optimizer.}. 
The parallel training capability of the PPO algorithm accelerates the learning process. 
Hyperparameter tuning is conducted to select parameters such as $\epsilon$, $\gamma$, $\lambda$, and Adam optimizer learning rate.

\section{Experiment}
\subsection{Scenario Setting}

\begin{figure}[thpb]
      \centering
      \framebox{\parbox{3.3in}{
      \centering
      \includegraphics[width=3.3in]{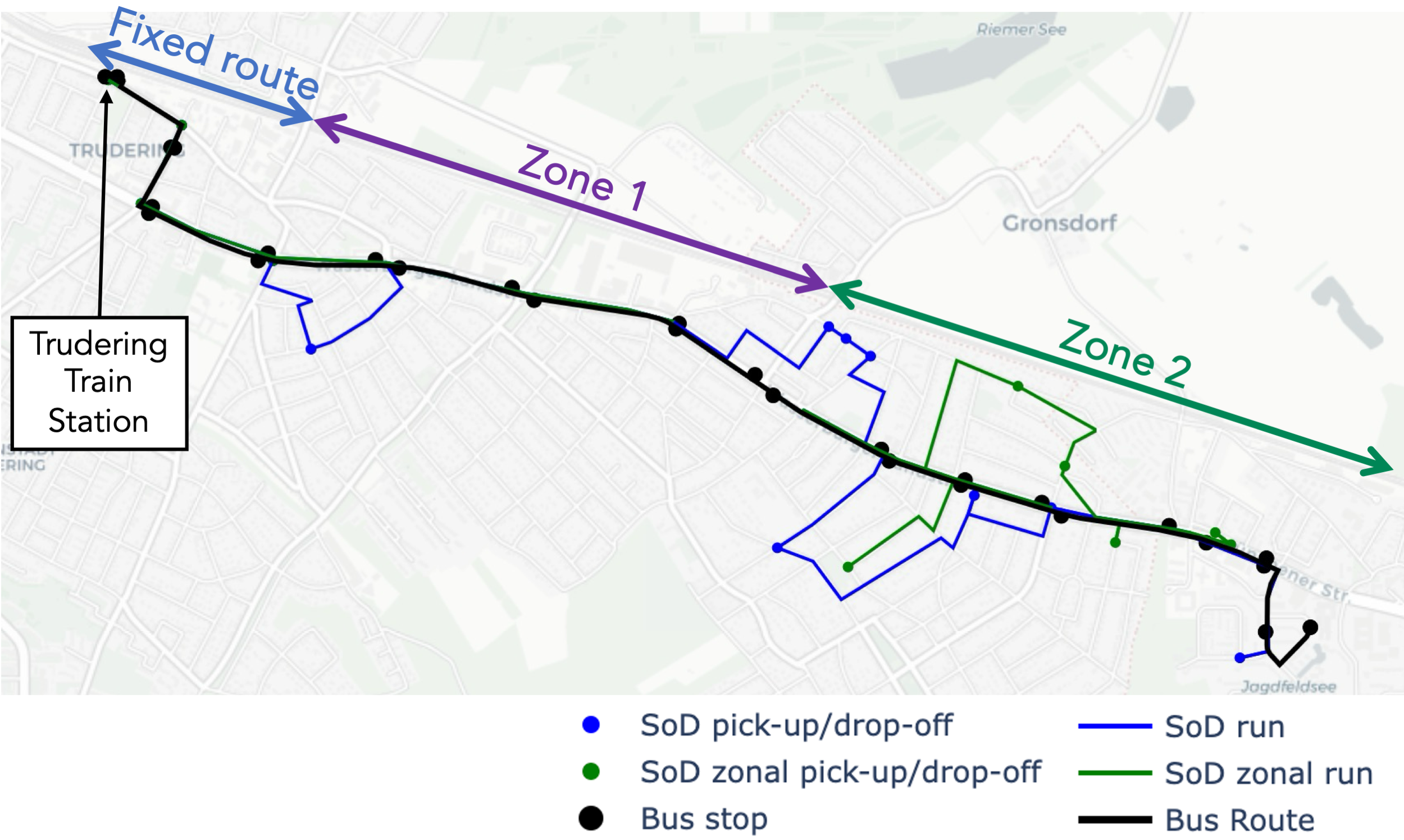}
      }}
    \caption{Sample trajectories of SAVs}
    \label{fig:trajectory}
    \end{figure}

To demonstrate the simulation with zonal dispatching control, bus route 193 in Munich, Germany, a feeder to the train station Trudering Bahnhof, is selected as an example.
Its route length is 5.6~km with a round-trip journey time of 33~min. 
The fixed route is set as the first 1.2~km, followed by two flexible zones of 2.2~km each.
\footnote{Refer to Ng et al. \cite{ng_autonomous_2023} for zone size analysis.}
The SAV trajectories in the simulation are illustrated in Fig.~\ref{fig:trajectory}, with a regular SoD route (blue) and a zonal SoD route to Zone 2 (green).

Simulation runs from 9~p.m. to midnight for off-peak demand, with the first hour ignored in metric evaluation for warm-up. 
Trip origins and destinations are adapted from a boarding and alighting dataset of a local public transit operator, whereas demand is assumed to be a linear function of walking time (maximum 10~min). 
GIS dataset of transit alignment and stop locations \cite{mvg_mvg_2024} and road network with steady travel time \cite{boeing_osmnx_2017} are integrated into the model.

Four control types are simulated and compared --- 1) fixed route; 2) SoD; 3) Zonal (SoD with nominal zonal dispatching); and 4) RL Zonal (SoD with RL-based zonal dispatching).
Eight SAVs are available.
For the first two control types, SAVs are dispatched to cover the whole route regularly at a 5-min headway.
For the last two, four SAVs are dedicated to regular SoD routes at a 10-min headway, with the remaining four sent to two zones regularly for type 3 (Zonal) and based on the RL model for type 4 (Zonal RL).

Other simulation parameters are as follows: $\gamma^A=\$33/h$, $\gamma^O=\$0.694/km$, $\gamma^R=\gamma^S=10^6$, $\gamma^T=\$16.5/h$, $\gamma^V=\$7.59/h$, $\gamma^W=\$24.75/h$, $\phi=2$, $\phi^P=2.5$, $c^V=20$, $t^{D,f}=300s$, $t^R=1200s$, $t^S=60s$, $t^{W,max}=900s$.\footnote{Monetary values are in US dollars. See \cite{tirachini_economics_2020} for SAV cost parameters.}

The RL model is trained on 2000 demand instances with eight parallel threads, i.e., 16\,000 simulations. 
Each simulation of 3~h consists of 180 times steps ($T=180$), totaling 2\,880\,000 time steps. 
Other RL parameters are as follows: $\epsilon=0.2$, $\gamma=0.99$, $\lambda=0.95$, batch size=$64$, epoch=$10$, learning rate=$0.003$.

The results of each control type are from simulations on 100 demand instances (separate from the RL training set). 
The RL results are from static deployment of the actor model (no on-policy learning) after training.

The RL model training and simulation were performed on a computer with Windows 11 operating system, Intel Xeon Silver 4214R 2.4GHz (12 cores), NVIDIA RTX A4000 (6144 CUDA cores), and 96~GB RAM. 
The PPO implementation \cite{stable-baselines3} is integrated with the FleetPy environment on Python 3.12.1.

\subsection{Results}

\begin{figure}[!thb]
      \centering
      \framebox{\parbox{3in}{
      \centering
      \includegraphics[width=3in]{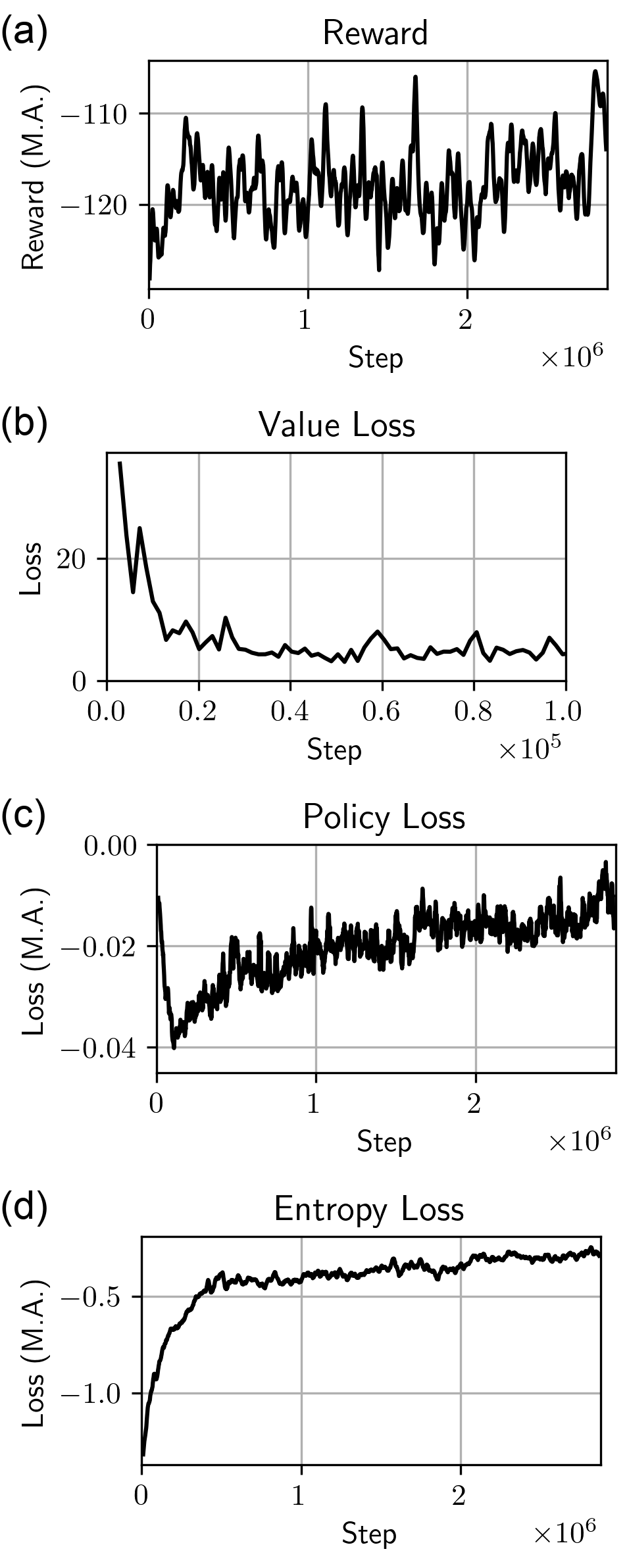}
      }}
    \caption{Training metrics of the RL model (M.A.: 10-step moving average)}
    \label{fig:rl_train}
    \end{figure}

The RL model training with over 2 million steps is completed within 24 hours (around 34 time steps per second).
The reward sees continuous improvement throughout the training, from worse than -130 initially to around -115 approaching the end in Fig.~\ref{fig:rl_train}a.
It fluctuates as the actor model explores policies. 
The critic model is trained fast with value loss falling below 5\% of the reward within the first 50\,000 steps (Fig.~\ref{fig:rl_train}b). 
This supports the actor model to improve the policy and reduce the policy loss magnitude gradually (Fig.~\ref{fig:rl_train}c). 
The balance of exploration and exploitation is seen in Fig.~\ref{fig:rl_train}d where the decreasing entropy indicates increasing stability of the policy. 

\begin{figure}[thpb]
      \centering
      \framebox{\parbox{3in}{
      \centering
      \includegraphics[width=3in]{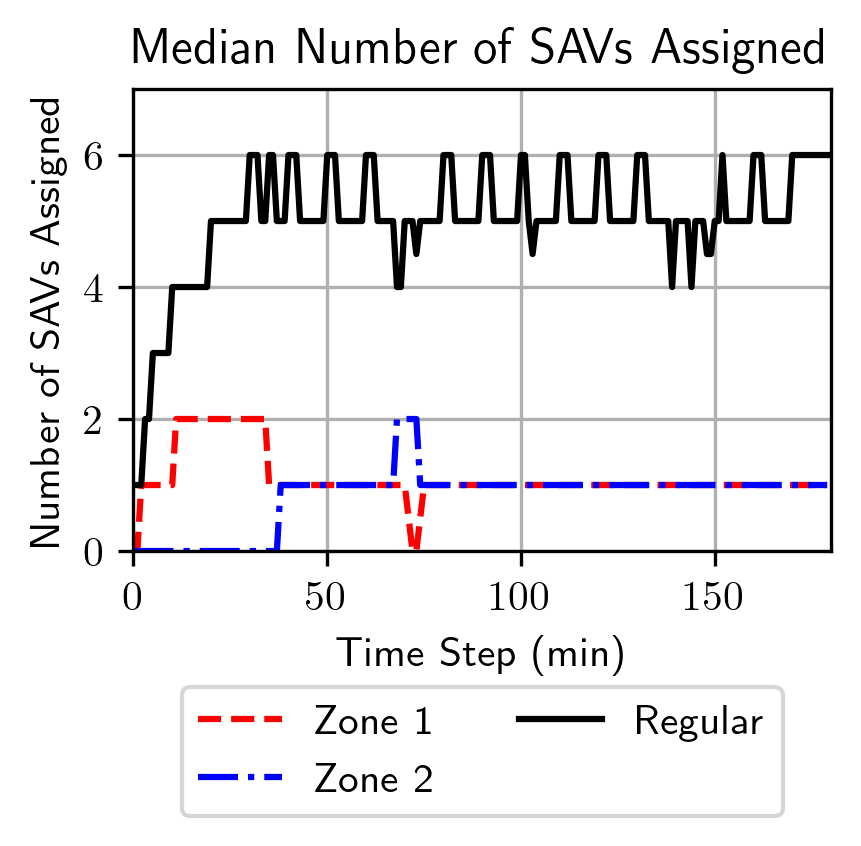}
      }}
    \caption{Number of SAVs assigned to each zone (95\% confidence interval shaded)}
    \label{fig:n_SAV}
    \end{figure}

\begin{figure}[thpb]
      \centering
      \framebox{\parbox{3in}{
      \centering
      \includegraphics[width=3in]{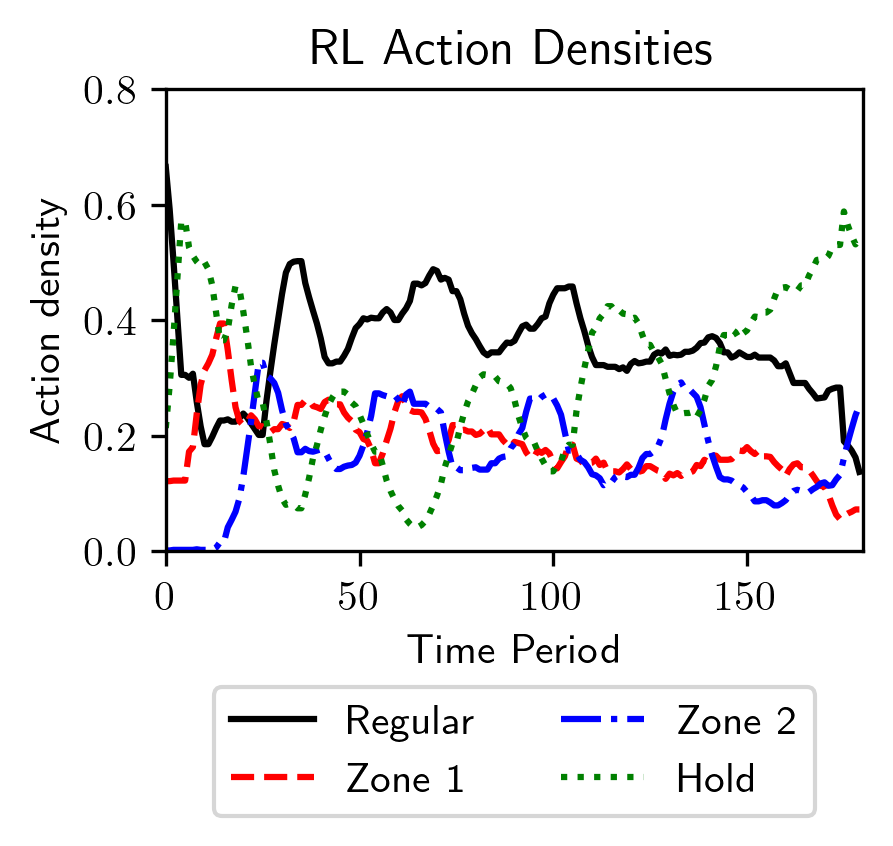}
      }}
    \caption{Densities of RL model actions}
    \label{fig:action_density}
    \end{figure}

The following results are from the 100 simulations run with the trained RL model.
The actor model suggests different actions based on demand\footnote{Demand falls gradually from 9~p.m. to midnight for the tested bus route.} and current SAV operations.
Fig.~\ref{fig:n_SAV} shows more SAVs assigned to regular routes.
Apart from the four SAVs reserved for regular SoD routes that are dispatched every 10~min, the RL model also assigns around two other vehicles to regular routes. 
The remaining SAVs serve Zone 1 first and then to both zones later. 
We also observe that the model sometimes strategically holds SAVs to avoid bunching and cater to future demand.
This is more obvious in Fig.~\ref{fig:action_density}, which shows the proportion of actions taken in different time steps across simulations.

\begin{figure}[thpb]
      \centering
      \framebox{\parbox{3in}{
      \centering
      \includegraphics[width=3in]{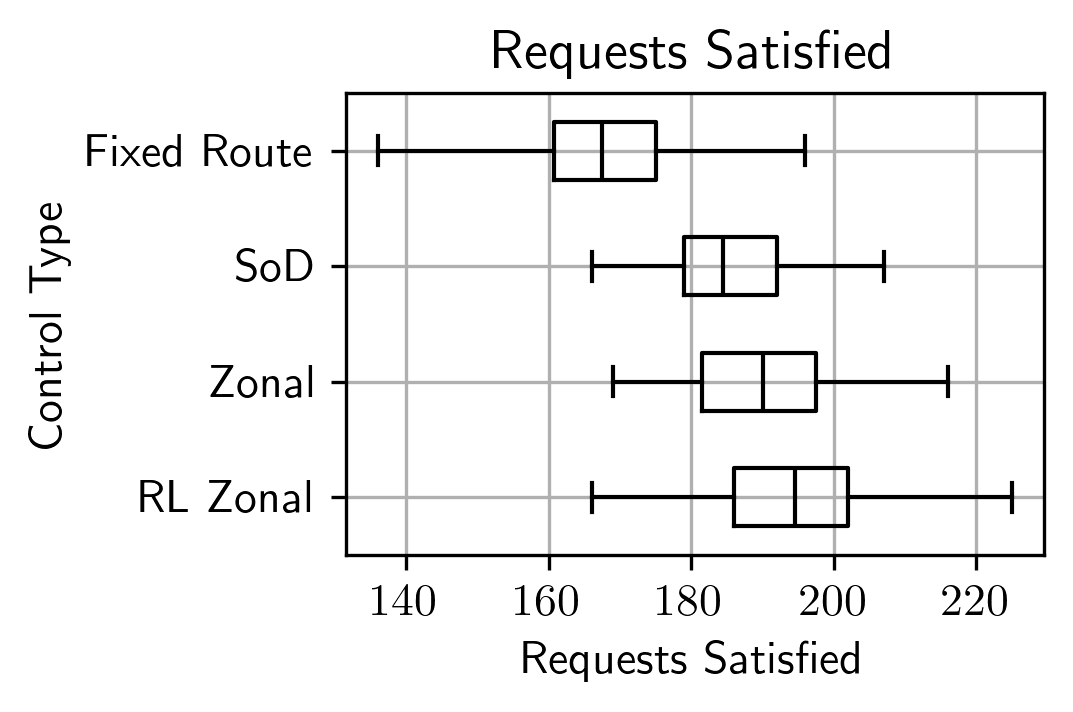}
      }}
    \caption{Requests satisfied of four control types}
    \label{fig:res_request}
    \end{figure}

\begin{figure}[thpb]
      \centering
      \framebox{\parbox{3in}{
      \centering
      \includegraphics[width=3in]{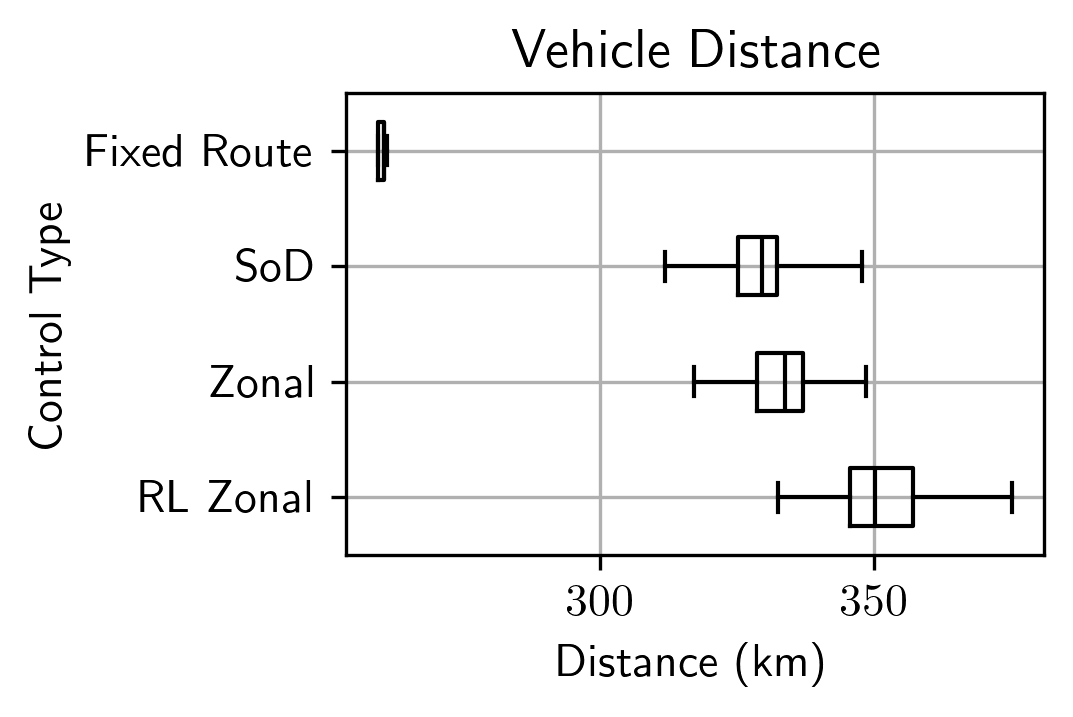}
      }}
    \caption{Vehicle distance traveled of four control types}
    \label{fig:res_dist}
    \end{figure}

The box plot in Fig.~\ref{fig:res_request} compares the results of the four service types, each from 100 simulations.\footnote{Boxplots show the minimum, first quartile, median, third quartile, and maximum.}
The convenience of SoD service to pick up and drop off passengers attracts more demand, resulting in 10\% more passengers served on average compared to the fixed route. 
Zonal RL control brings higher efficiency and serves 16\% more passengers, or 2.4\% more compared to the nominal zonal service. 
Serving more passengers naturally comes at a cost of higher vehicle miles traveled (Fig.~\ref{fig:res_dist}), due to more flexible requests served and additional trips made.
    
\begin{figure}[!thb]
      \centering
      \framebox{\parbox{3in}{
      \centering
      \includegraphics[width=3in]{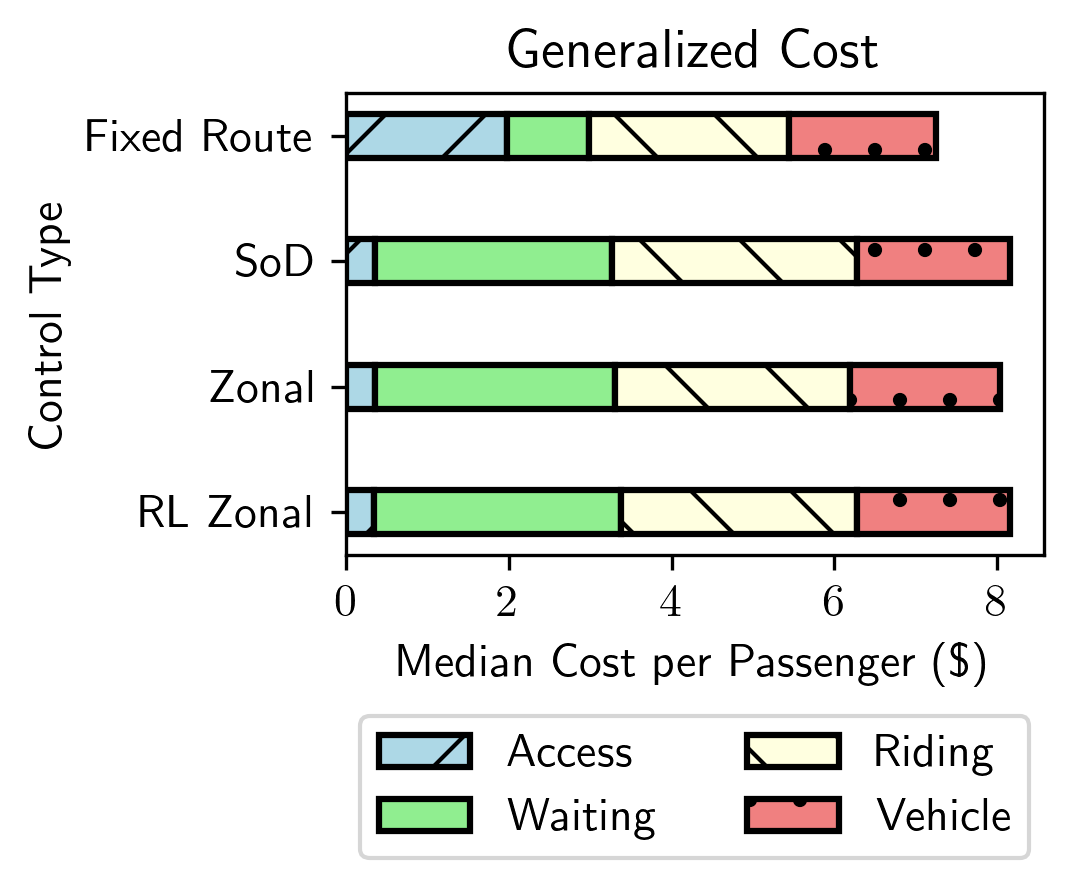}
      }}
    \caption{Generalized costs of four control types}
    \label{fig:res_cost}
    \end{figure}

Fig.~\ref{fig:res_cost} breaks down the generalized costs per passenger for each service type. 
SoD services reduce passengers' access time considerably (entirely in the flexible route portion). 
However, this is accompanied by increases in waiting and riding times due to detours and headway variances.
The overall generalized cost increases by 13\% from fixed route to SoD.
Nevertheless, the costs of RL zonal control are similar to those of SoD (within 0.1\%).
This suggests that RL zonal control serves more passengers with a similar cost, thanks to the efficiency improvement.

\section{Conclusions}

In this paper, we have introduced a novel semi-on-demand (SoD) transit feeder service leveraging shared autonomous vehicles (SAVs) and a reinforcement learning (RL)-based zonal dispatching control strategy. 
The service combines the economy of scales of fixed routes and flexibility of on-demand services.
This offers a promising solution to the first-mile-last-mile problem, particularly in scenarios of directional demand, such as transit feeders.

The RL dispatching control, supported by the policy gradient method Proximal Policy Optimization, has proven effective in dynamically aligning vehicle zonal dispatching with fluctuating demand and operations, thus maximizing service coverage without excessive detours or waiting times. 
The framework with the FleetPy environment simulates SAV operations and assesses the performance of different dispatching strategies.

The experimental results from simulations using a Munich bus route demonstrate an increase in passengers served without considerable increases in user's and operator's costs. 
The RL zonal dispatching outperforms traditional fixed-route and SoD services in terms of operational efficiency and service flexibility.

Built on the SoD service concept and RL control of this study, future research could address some limitations for more in-depth insights. 
For example, the RL reward function could consider generalized costs or other comprehensive metrics, further improving the economic viability and user experience. 
More sophisticated control to adjust the number of zones, zone boundaries, and schedules could be experimented with RL.
The higher demand brought by convenient SoD service may justify deploying more SAVs and further improving the service, reducing waiting time in particular.
The benefits of RL in demand fluctuation could also be tested for further study in resilience. 
Lastly, the simulation could be further expanded to incorporate dynamic travel times, multimodal demand models, and more network types.

In summary, this study paves the way for future innovations in public transit, particularly through the integration of SAV feeders and machine learning methods. 
Such technologies could transform public transit networks into more adaptive, efficient, and user-focused systems, contributing to sustainable urban mobility.

\bibliographystyle{IEEEtran}
\bibliography{IEEEabrv,references}

\end{document}